\documentclass[conference]{IEEEtran}
\IEEEoverridecommandlockouts
\usepackage{cite}
\usepackage{amsmath,amssymb,amsfonts}
\usepackage{algorithmic}
\usepackage{graphicx}
\usepackage{textcomp}
\usepackage{xcolor}
\usepackage{caption}
\def\BibTeX{{\rm B\kern-.05em{\sc i\kern-.025em b}\kern-.08em
    T\kern-.1667em\lower.7ex\hbox{E}\kern-.125emX}}
\begin{document}

\title{A Medical Pre-Diagnosis System for Histopathological Image of Breast Cancer\\
}

\author{\IEEEauthorblockN{Shiyu Fan\textsuperscript{1,*,a,\dag}}
\IEEEauthorblockA{$^{1}$Department of Computer Science\\
George Washington University\\
Washington, D.C., United States\\
$^{a}$sfan47@gwu.edu}
\and
\IEEEauthorblockN{Runhai Xu\textsuperscript{2,b,\dag}}
\IEEEauthorblockA{$^{2}$Department of Mathematics\\
University of Waterloo\\
Waterloo, Canada\\
$^{b}$r2999xu@uwaterloo.ca}
\and
\IEEEauthorblockN{Zhaohang Yan\textsuperscript{3,c,\dag}}
\IEEEauthorblockA{$^{3}$Department of Computer \&\\
Mathematical Sciences\\
University of Toronto\\
Toronto, Canada \\
$^{c}$zhaohang.yan@mail.utoronto.ca\\
$^{\dag}$These authors contribute equally.}
}
\maketitle

\begin{abstract}
This paper constructs a novel intelligent medical diagnosis system, which can realize automatic communication and breast cancer pathological image recognition. This system contains two main parts, including a pre-training chatbot called M-Chatbot and an improved neural network model of EfficientNetV2-S named EfficientNetV2-SA, in which the activation function in top layers is replaced by ACON-C. Using information retrieval mechanism, M-Chatbot instructs patients to send breast pathological image to EfficientNetV2-SA network, and then the classifier trained by transfer learning will return the diagnosis results. We verify the performance of our chatbot and classification on the extrinsic metrics and BreaKHis dataset, respectively. The task completion rate of M-Chatbot reached 63.33\%. For the BreaKHis dataset, the highest accuracy of EfficientNetV2-SA network have achieved 84.71\%. All these experimental results illustrate that the proposed model can improve the accuracy performance of image recognition and our new intelligent medical diagnosis system is successful and efficient in providing automatic diagnosis of breast cancer.
\end{abstract}


\section{Introduction}
Breast cancer is a kind of malignant tumor disease that occurs frequently in women. For the potential patients, pathological analysis of breast tissue is considered as a mature and diagnostic method for breast cancer. Early diagnosis of malignant tissue can effectively help patients receive the correct treatment instantly. With the improvement of automatic algorithms for pathological images classification, the Computer-Aids Diagnosis (CAD) system has become an auxiliary tool and has been commonly applied in pathological analysis.

To better analyze the pathology, chatbot is increasingly applied in the healthcare industry, which can help doctors to perform information extraction and offer patients professional medical assistance. The benefits that chatbot brings to CAD systems are significant. It can realize 24-hour real-time operation without the need for professionals to be on the sidelines. With the development of machine learning, chatbot can respond to more topics and generate more anthropomorphic sentences.

Meanwhile, the medical profession's requirements for computerized medical image recognition are also increasing. At present, the classification accuracy of the non-neural network, mostly used by the CAD systems in breast cancer classification tasks, is not satisfactory. This kind of system cannot be used as a pre-diagnosis system for patients without professional knowledge. Due to the robust feature extracting ability, in recent years, researchers have attempted to use convolutional neural networks (CNNs) as the classifier of pathological images in the medical diagnosis system \cite{b1}. Afterward, CNNs have become mainstream of image classification algorithm for breast cancer.

Both chatbot and CNNs are widely utilized in the service consultation and medical image diagnosis domains. Meanwhile, with the explosive development of medical intelligent diagnosis and treatment systems, diagnosis for breast cancer pathological image is concerned as a significant part. Patients expect effective diagnosis in the absence of professional doctors. However, there is no precedent for combining these two techniques before. Under such circumstances, the integration of chatbot and pathological image recognition based on CNN is of great value.

To solve the problems that people may face in the process of self-diagnose for breast cancer, we propose an intelligent medical pre-diagnosis framework, which consists of two components:
\begin{itemize}
\item \textbf{M-ChatBot} for communicating with patients. For these non-professionals, M-Chatbot aims to provide professional guidance and assist them complete the imaging diagnosis process.
\item \textbf{EfficientNetV2-SA} for classifying breast cancer pathologic images. Given the segmented image, the mammography network aims to identify the type of pathological section tissue and return confidence of this diagnostic result to users.
\end{itemize}

Our system is verified on BreaKHis dataset \cite{b2}. The test results indicate that our brand-new system can understand the user’s sentences and give the corresponding responses correctively. Additionally, it can generate convincing results of the pathological diagnosis. By using our new system, patients can get a quick pre-diagnosis independently.

\section{Related Work}
As hotspots, in recent years, researchers have proposed many automatic dialog systems and image classification methods based on machine learning.

Chatbot has been a research focus in the field of natural language processing since Eliza was invented in 1966 \cite{b3}. In recent years, chatbot has two main directions, retrieval based chatbot and generative chatbot. Retrieval-based chatbot relies on a large number of dialogue databases which can provide predefined responses \cite{b4}. Nowadays, chatbots have been used in medical information services and disease diagnosis in view of the development of chatbot \cite{b5}.

In terms of image classification, methods applied for pathological image recognition in CAD systems are divided into two main implementation methods, Bag-of-words (BoW) model \cite{b6} and neural network \cite{b7}. The traditional BoW model has always been the mainstream algorithm before AlexNet \cite{b8} was proposed. The researchers prove that the classifier using the BoW model can achieve satisfactory results in x-ray data set \cite{b9}. As a rising star, deep neural networks have undergone multiple generations of evolution, and they have now been used in the task of classifying lung image patches with interstitial lung disease (ILD) \cite{b10}. As the updated network,  the EfficientNetV2 \cite{b11} proposed in 2021 simplifies the convolution module on the basis of EfficientNet \cite{b12} to cut down the training time. It also applies an improved progressive learning method, which dynamically adjusts the regularization method according to the size of the training image.

\section{Methodology}
This section describes the fundamental method of our medical pre-diagnosis system. Figure 1 illustrates the architecture of this proposed system. As shown in Figure 1, our system is divided into three modules, user’s input, chatbot and classifier. The user can communicate with the chatbot and upload the image to the classifier for pathological analysis. Section A and Section B provide details of our medical chatbot and pathological image classification network.

\begin{figure}[htbp]
\centerline{\includegraphics[width=0.48\textwidth]{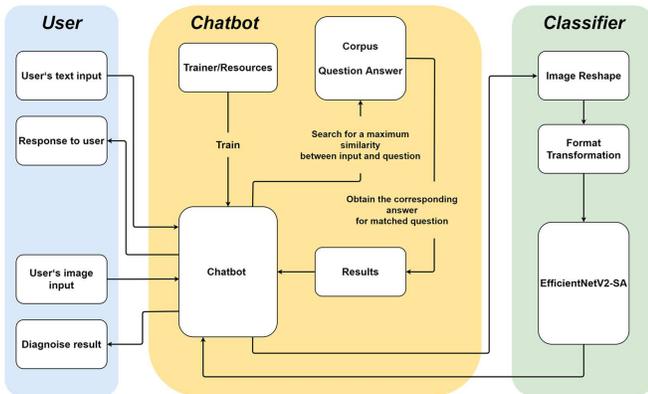}}
\caption{The architecture of our medical pre-diagnosis system framework.}
\label{fig1}
\end{figure}

\subsection{M-ChatBot}\label{AA}
The structure of our M-ChatBot mainly includes three parts through the implementation, storage adapter, logic adapter and trainer.

The storage adapter provides an interface that allows it to build connections with the storage which provides the functionality of text-searching. For example, if the sentence “Google, is. the best searching engine in the World” needs to find a matching. Then, the text-searching involves the following steps:
\paragraph{Capitals and Spacing are normalized}
“google is the best searching engine in the world”
\paragraph{Remove stop word}
“google best searching engine world”
\paragraph{Stem text for each word}
“oogl es earchin ngin orl”
\paragraph{Generate bigrams for each neighboring word}
“oogles esearchin earchinngin nginorl”

Bigrams are generated in the last step since it makes sorting, categorization and retrieval feasible even in a large set of documents \cite{b13}.

The number of matches is smaller compared to unigram since it is based on the probability of finding two neighboring words in an existing string that can be a match to our input.

The logic adapters decide what kinds of logic that M-ChatBot is used to select a response for a given input. Typically, the logic adapters will achieve this using two main steps. The first step is to search the database that closely matches the input statement. There are several methods that can be used to find a match. M-Chatbot tries to find a match with the minimum levenshtein distance to our input, meaning that there exists a maximum similarity between them since levenshtein distance is the minimum number of editing to transform one text to another \cite{b14}. Similarity calculation formula is as follows:
\begin{equation}
\mathit{Similarity}= 1-\frac{\mathit{levenstein\:distance}}{\max{Len}}
\end{equation}
Where $\max{Len}$ is the maximum string length of the text before and after the transformation \cite{b14}.

Before the matching process, a maximum similarity threshold is enforced, indicating the maximum amount of similarity between two statements that is required. The searching process continues until a match is found with a similarity calculated greater or equal to the threshold value. Once the match is selected, the second step then involves selecting a known response from the databases. Usually there can be several responses for the corresponding match, so the method of response selection also needs to be taken into consideration.

The training process of M-ChatBot involves loading prepared dialog examples into the databases. This creates a graph data structure which represents the set of statements and responses in the database. The trainer is provided with a dataset so that it can help build the M-Chatbot knowledge graph using provided statements and responses as entries so that all the information in the dialogue is clearly represented \cite{b15}. Figure 2 represents the combination of multiple dialogues for storage purpose.

\begin{figure}[htbp]
\centerline{\includegraphics[width=0.48\textwidth]{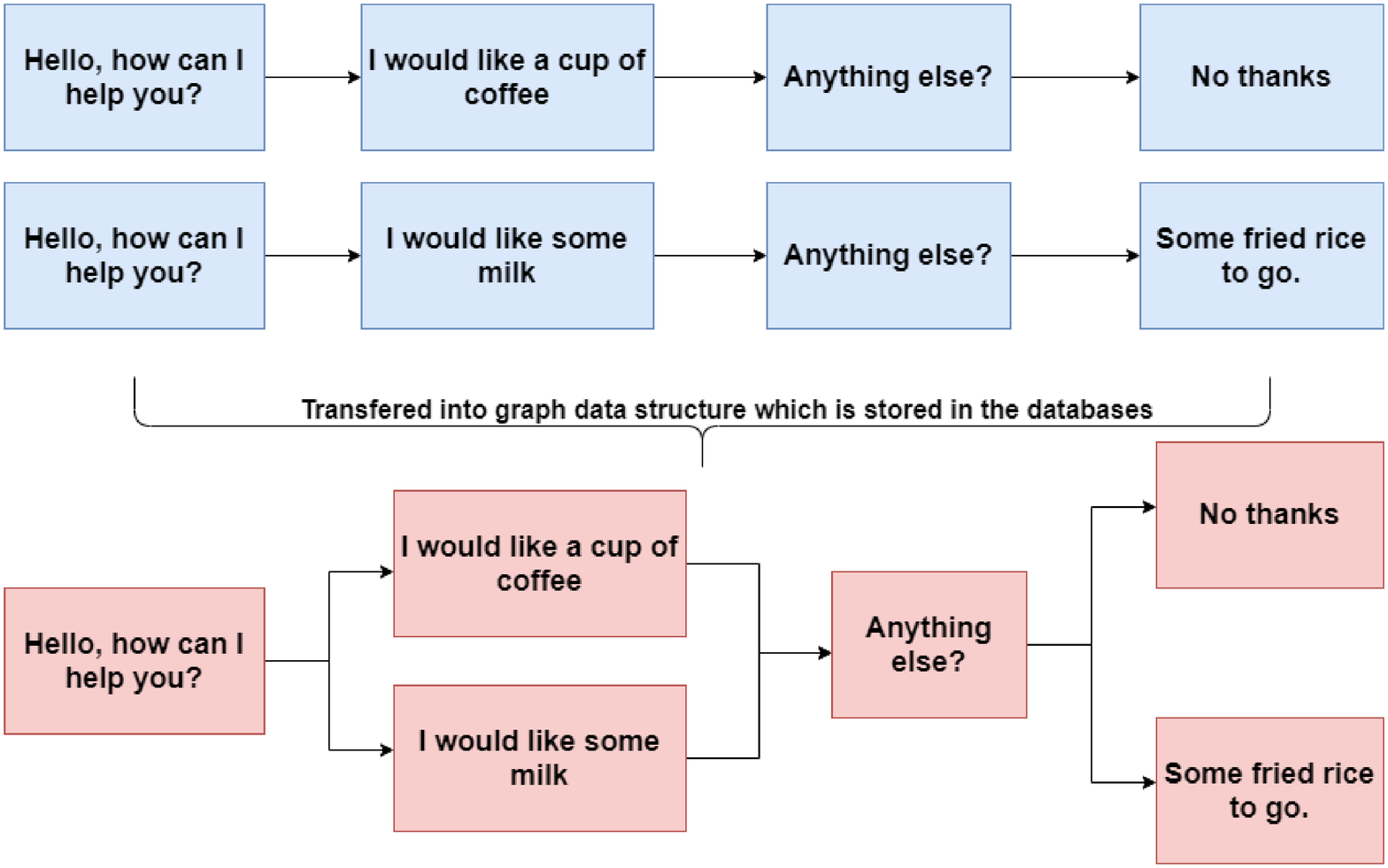}}
\caption{Combination of multiple dialogues for storage purpose.}
\label{fig2}
\end{figure}

\subsection{EfficientNetV2-SA}
For pathological image classification, we use EfficientNetV2-SA, which is modified from the EfficientNetV2-S model  and accelerate the training of the network through transfer learning.

The detailed structure of EfficientNetV2-SA is shown in Table 1. The image feature extraction part is retained and the activition function of  Dense layer is replaced by ACON-C function \cite{b16}.

In order to save the amount of parameters while improving accuracy, EfficientNetV2-SA’s feature extraction module is consistent with EfficientNetV2-S’ convelutional layers. It includes of three Fused-MBConv layers and three MBConv layers. Comparing to the MBConv layer, the Fused-MBConv abandons the combination of a Depwise Conv and a Cov1x1 and replace it with a traditional Conv 3x3 module. We also retain the incremental learning method originated from EfficientNetV2.

As shown in the Fig3, the trainable top layers of our model has no change except the activation. One Conv1x1 layer, one global average pooling(GAP) layer \cite{b17} and one dense layer constitudes this feature classification network of this model. GAP layer can caculate the average of each feature map and send the resulting vector to the softmax layer. In the task of transfer learning, using a GAP layer instead of a fully connected layer can reduce the amount of parametors and avoid overfitting.

The main improvement of this model is the introduction of ACON-C activation. ACON-C is defined as follows:
\begin{equation}
f_{\mathrm{ACON-C}}(x)=S_{\beta }(p_{1}x,p_{2}x)=(p_{1}-p_{2})x\sigma [\beta (p_{1}-p_{2})x]+p_{2}x
\end{equation}
where $p_{1}$ and $p_{2}$ are two trainale parameter and $\sigma$ is denoted as Sigmoid function. The most popular activation - Swish function \cite{b18} can be considered as a smooth approximation of ReLU \cite{b19}. It takes advantage in very deep netwrkdue to it’s smooth and non-monotonic characteristics. The Swish function is a special member of the ACON family. ACON-C is a general activation function which is a smooth approximation of Maxout \cite{b20} family(e.g. Leaky ReLU \cite{b21}, PReLU \cite{b22}, etc).

\begin{table}[htbp]
\caption{Detailed configuration of EfficientNetV2-SA}
\begin{center}
\begin{tabular}{ccccc}
\hline
\textit{\textbf{Stage}} & \textit{\textbf{Operator}} & \textit{\textbf{Channels}} & \textit{\textbf{Activation}} & \textit{\textbf{Layers}} \\ \hline
0                       & Conv 3x3                   & 24                         & SiLU                         & 1                        \\
1                       & Fused-MBConv1, k3x3        & 24                         & SiLU                         & 2                        \\
2                       & Fused-MBConv4, k3x3        & 48                         & SiLU                         & 4                        \\
3                       & Fused-MBConv4, k3x3        & 64                         & SiLU                         & 4                        \\
4                       & MBConv4, k3x3, SE0.25      & 128                        & SiLU/Sigmoid                 & 6                        \\
5                       & MBConv6, k3x3, SE0.25      & 160                        & SiLU/Sigmoid                 & 9                        \\
6                       & MBConv6, k3x3, SE0.25      & 272                        & SiLU/Sigmoid                 & 15                       \\
7                       & Conv 1x1, BN               & 272                        & ACON-C                       & 1                        \\
8                       & Pooling                    & 1792                       &                              & 1                        \\
9                       & Dense                      & 1792                       &                              & 1                        \\ \hline
\end{tabular}
\label{tab1}
\end{center}
\end{table}

\begin{figure}[htbp]
\centerline{\includegraphics[width=0.48\textwidth]{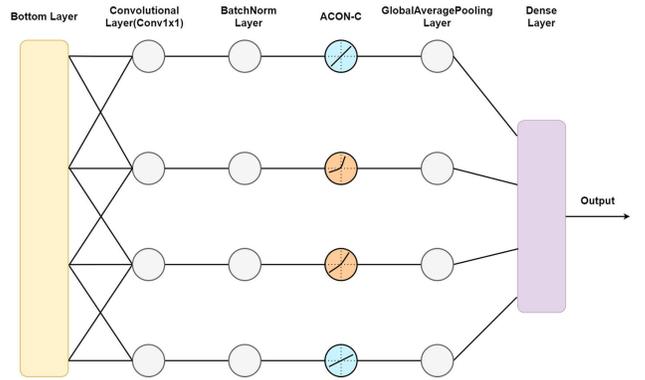}}
\caption{The structure of EfficientNetV2-SA' top layers.}
\label{fig3}
\end{figure}

\section{Experiment Settings}
This section introduces the detailed experimental settings. According to the framework of our system, the experiments focus on two aspects-the dialogue and classification performances.

\subsection{Implementation Details}
\paragraph{M-ChatBot}In our chatbot, the maximum similarity threshold is equal to 0.90, and the default response is set to be “-I am sorry, but I do not understand” when it is unable to get a match with a sentence stepping across our threshold. At last, the trainer is set to be a list trainer, which allows a chat bot to be trained using a list of strings where each element in the list indicates a sentence belonging to the conversation so that the whole list forms a complete conversation.

There is a function implemented specifically for the training process. It takes a list which consists of all file names, since there aren’t many online resources closely related to breast diagnosis conversations. To provide sample conversations, there are multiple files made for the training process. For example, “greetings.txt” for some basic greeting conversations, “doctor.txt” is made to simulate the process of inquiry since it contains sample conversations between a doctor and a patient. This function reads through the list and grasps all the files that are needed in the training procedure and feeds the chatterbot line by line through the training process so that it absorbs all the information included in those files whose names appear on the list.

When the main function gets called, the medical bot can start chatting to patients from any topics. Meanwhile, to give patients a basic understanding of how likely they are to get breast cancer, our chatbot will store the details of condition from the dialogue and do some analysis based on those details. If the probability of getting breast cancer is very high based on these words given by the patient, we need to do some further provided, which is, of course, more accurate and credible.
\paragraph{EfficientNetV2-SA}We compare the performance of EfficientNetV2-SA network with VGG-16 \cite{b23}, ResNet-50 \cite{b24} and EfficientNet-S.The trainable layer structure of these five network models is listed in Table II. All the pretrained models are based on ImageNet ILSVRC2012 classification weights \cite{b25}.

To train the EfficientNetV2-SA, we use TensorFlow 2.4 as implementation and Adam optimization algorithm \cite{b26} as optimizer. We set learning-rate to 0.001, batch size to 16 and epochs to 200. To get better generalization performance, we apply the early stopping mechanism. The training setting of the remaining models are also consistent with it.

\begin{table*}[htbp]
\centering
\setlength{\tabcolsep}{6mm}{
\caption{The top layers' structures of five models}
\begin{tabular}{ccccc}
\hline
\textit{\textbf{VGG-16FC}} & \textit{\textbf{VGG-16GAP}} & \textit{\textbf{ResNet-50}} & \textit{\textbf{EfficientNetV2-S}} & \textit{\textbf{EfficientNetV2-SA}} \\ \hline
Maxpooling                 & Conv1x1                     & Conv1x1                     & Conv1x1                            & Conv1x1                             \\
FC-1024                    & BN(SiLU)                    & BN(SiLU)                    & BN(SiLU)                           & BN(ACON-C)                          \\
Dropout 0.3                & Averagepooling              & Averagepooling              & Averagepooling                     & Averagepooling                      \\
FC-512                     & softmax                     & softmax                     & softmax                            & softmax                             \\
softmax                    & \textbackslash{}            & \textbackslash{}            & \textbackslash{}                   & \textbackslash{}                    \\ \hline
\end{tabular}}
\end{table*}
\subsection{Dialogue Test Settings}
According to the statistics, the metrics of medical chatbot studies are diverse with survey designs and global usability metrics dominating \cite{b26}. Since our system focuses on certain tasks, we evaluated the performance of M-Chatbot with three measurement metrics, comprehension capability, goal completion rate (GCR) and user’s satisfaction score. For GCR and user’s satisfaction, we randomly invited 80 statistics students who are proficient in English. Everyone was asked to complete 5 rounds of dialogue and record GCR and the user's satisfaction with the dialogue.

\subsection{Image Dataset Selection and Data Preprocessing}
We train and test our model on BreaKHis breast pathology image dataset. BreaKHis totally contains 7909 marked breast pathology images, of which 2480 are benign and 200 are malignant. The dataset contain four distinct types of benign breast tumors: adenosis (A), fibroadenoma (F), phyllodes tumor (PT) and tubular adenoma (TA); and four malignant tumors : ductal carcinoma (DC), lobular carcinoma (LC), mucinous carcinoma (MC) and papillary carcinoma (PC). Each image has four magnifications, 40X, 100X, 200X and 400X. All images are in 720x360 pixel RGB format. The details of BreakHis are shown in Table II.

\begin{table}[htbp]
\caption{Benign and malignant distribution in BreaKHis}
\begin{center}
\begin{tabular}{lccc}
\hline
\textit{\textbf{Magnification}} & \multicolumn{1}{l}{\textit{\textbf{Benign}}} & \multicolumn{1}{l}{\textit{\textbf{Malignant}}} & \multicolumn{1}{l}{\textit{\textbf{Total}}} \\ \hline
\textit{\textbf{40X}}           & 625                                          & 1370                                            & 1995                                        \\
\textit{\textbf{100X}}          & 644                                          & 1437                                            & 2081                                        \\
\textit{\textbf{200X}}          & 623                                          & 1390                                            & 2013                                        \\
\textit{\textbf{400X}}          & 588                                          & 1232                                            & 1820                                        \\
\textit{\textbf{Total}}         & 2480                                         & 5429                                            & 7909                                        \\ \hline
\textit{\textbf{Patients}}         & 24                                           & 58                                              & 82                                         
\end{tabular}
\end{center}
\end{table}

In our experiment,BreaKHis dataset is randomly divided into training (70\%) and testing (30\%) set. All images have not undergone additional enhancement processing.

\begin{figure}[htbp]
\centerline{\includegraphics[width=0.48\textwidth]{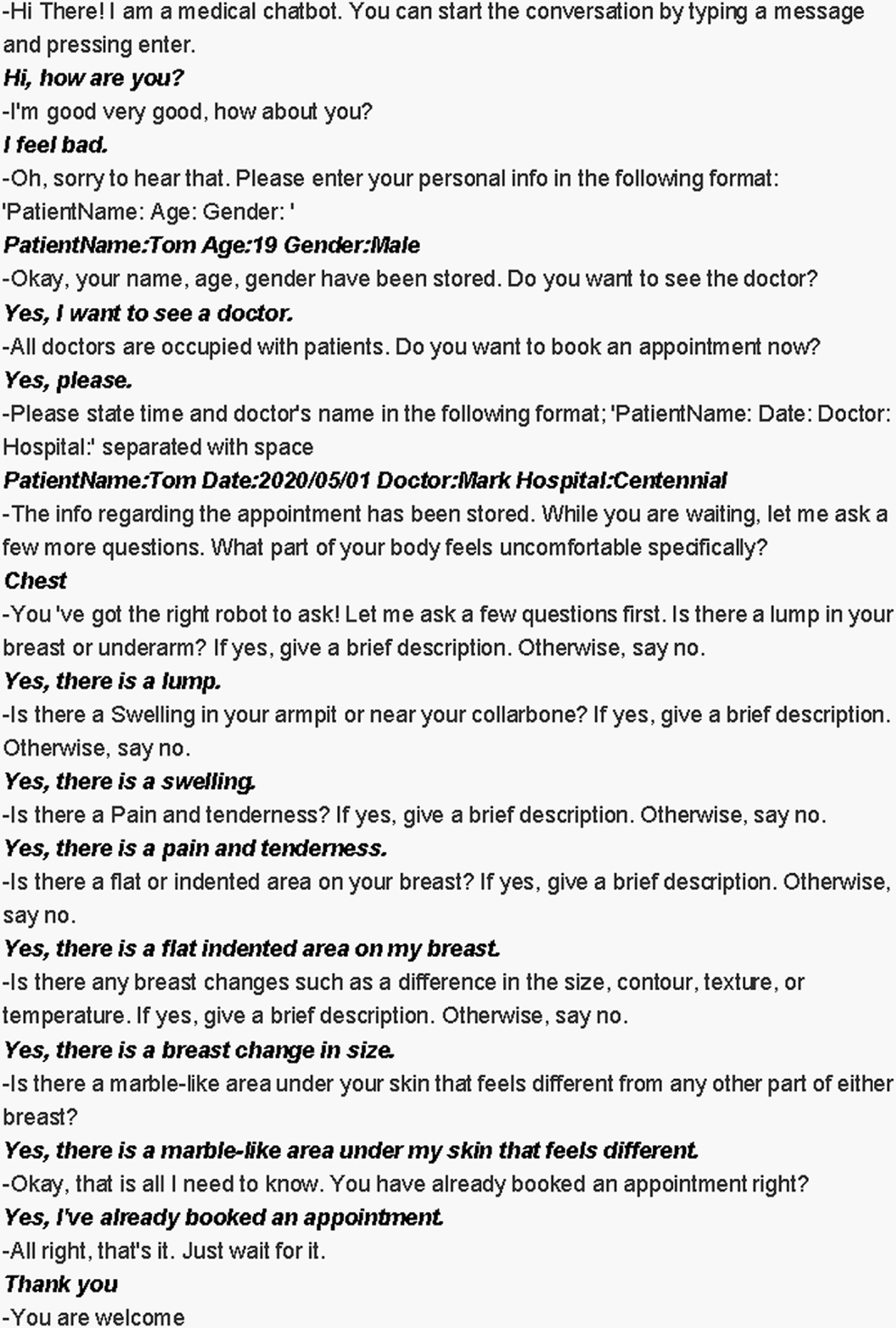}}
\caption{Example of a successful dialogue.}
\label{fig4}
\end{figure}

\begin{figure}[htbp]
\centerline{\includegraphics[width=0.48\textwidth]{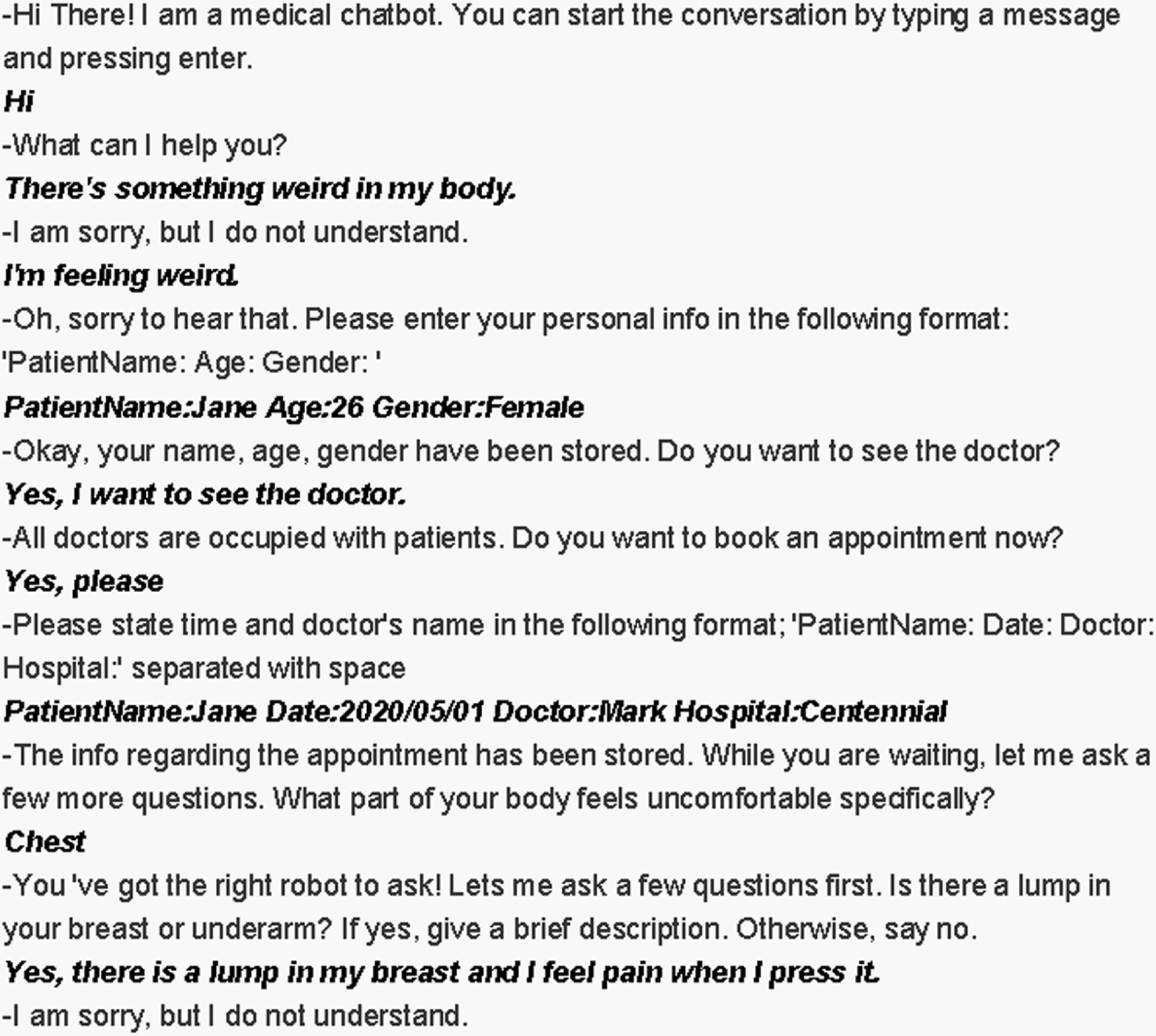}}
\caption{Example of a non-response dialogue.}
\label{fig5}
\end{figure}

\section{Experimental Results}
\subsection{Comprehension Capability}
Since the threshold is set to 90\% so that it makes sure that the sentence provided by the user has a high probability to match a sentence in the training dataset, then the chatbot can confidently choose the corresponding response for the matched sentence as the answer to the user. Tests on the ability of auto correction regarding user’s typo is performed, with around 53.33\%, 8 out of 15 tests are passed, also due to the very high standard I set for a match, must be extremely similar (90\% threshold).

\subsection{Goal Completion Rate (GCR)}
The goal of the M-ChatBot is to gather potential breast cancer patients’ basic information by doing some fundamental medical inquiry virtually. Therefore, the rate of completing this specific goal successfully is an important evaluation of the M-ChatBot. Each dialogue can be labeled as “Completed” or “Failed” by analyzing whether it completes the goal or not [17]. In this way, we calculate the GCR. The result is shown in Table IV.

\begin{table}[htbp]
\caption{Performance evaluation of the M-ChatBot with respect to GCR}
\begin{center}
\begin{tabular}{p{2cm}p{2cm}p{2cm}}
\hline
\textit{\textbf{Number of testing dialogues}} & \textit{\textbf{Number of completed goals}} & \textit{\textbf{Goal Completion Rate}} \\ \hline
30                                            & 19                                          & 63.33\%                                \\ \hline
\end{tabular}
\end{center}
\end{table}

To show how M-ChatBot works clearly, we give an example of a dialogue that completed the goal successfully in the Figure 4 (the bold and italic text is the user input).

The M-ChatBot uses dialogue to obtain the information of the user. Doctors can have a rough initial diagnosis by looking at the information it gathered without spending time doing it by doctors themselves.

To show the remaining problems, we give an example of no response. Figure 5 is a dialogue that the M-Chatbot failed to respond as we desire (the bold and italic text is the user input).

From the example 2 we can see that in the last dialogue, the M-ChatBot could not respond to the user properly. The reason for this problem is that in order to improve the accuracy of the answer, the maximum similarity threshold is set too high.

\subsection{User’s Satisfaction}
In this part, participants in the survey need to evaluate the dialogue satisfaction after completing a round of dialogue. There are five levels of user satisfaction evaluation and each level has the following definition. The scoring criteria is shown in Table V and Fig.5 represents the survey result.

\begin{table}[htbp]
\caption{Performance evaluation of the M-ChatBot with respect to GCR}
\begin{center}
\begin{tabular}{ll}
\hline
\multicolumn{1}{c}{\textit{\textbf{Level}}} & \multicolumn{1}{c}{\textit{\textbf{Evaluation Standard}}} \\ \hline
5 points                                    & Complete the task successfully; No grammatical errors     \\
4 points                                    & Few grammatical errors                                    \\
3 points                                    & Wired; Barely complete the task                           \\
2 points                                    & Task failed; Lots of grammatical errors                   \\
1 point                                     & Completely wrong answer for the whole time                \\ \hline
\end{tabular}
\end{center}
\end{table}

\begin{figure}[htbp]
\centerline{\includegraphics[width=0.48\textwidth]{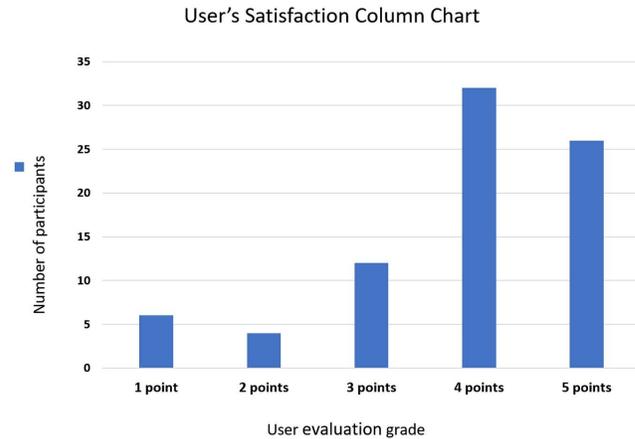}}
\caption{User satisfaction score distribution map.}
\label{fig6}
\end{figure}

\begin{table}[htbp]
\caption{Result of four classifier accuracy experiment}
\begin{center}
\begin{tabular}{lcccc}
\hline
\textit{\textbf{}} & \textit{\textbf{40X}} & \textit{\textbf{100X}} & \textit{\textbf{200X}} & \textit{\textbf{400X}} \\ \hline
VGG-16FC           & 75.9                  & 72.7                   & 70.5                   & 69.9                   \\
VGG-16GAP          & 82.4                  & 80.1                   & 80.7                   & 77.2                   \\
ResNet-50          & 84.4                  & 83.2                   & 81.2                   & 80.1                  \\
EfficientNetV2-S   & 83.3                  & 83.5                   & 82.1                   & 82.3                  \\
EfficientNetV2-SA  & 89.0                  & 83.9                   & 83.3                   & 82.7                   \\ \hline
\end{tabular}
\end{center}
\end{table}

\begin{table*}[ht]
\caption{Result of four classifier accuracy experiment}
\begin{center}
\setlength{\tabcolsep}{5mm}{
\begin{tabular}{cccccccccc}
\hline
\textit{\textbf{}}   & \textit{\textbf{}}   & \multicolumn{8}{c}{\textit{\textbf{Accuracy}}}                                                 \\ \hline
\multicolumn{1}{l}{} & \multicolumn{1}{l}{} & \multicolumn{4}{c}{\textit{\textbf{Benign}}} & \multicolumn{4}{c}{\textit{\textbf{Malignant}}} \\
                     &                      & A         & F         & PT        & TA       & DC         & LC         & MC        & PC        \\
40X                  & EfficientNetV2-S     & 81.3      & 82.9      & 83.8      & 90.9     & 83.1       & 80.9       & 79.1      & 81.4      \\
                     & EfficientNetV2-SA    & 82.4      & 89.3      & 84.4      & 100.0    & 91.5       & 87.0       & 81.9      & 90.7      \\
100X                 & EfficientNetV2-S     & 83.9      & 83.1      & 88.9      & 88.6     & 82.2       & 76.1       & 72.7      & 76.2      \\
                     & EfficientNetV2-SA    & 83.8      & 74.3      & 91.1      & 93.2     & 90.7       & 82.6       & 80.3      & 67.6      \\
200X                 & EfficientNetV2-S     & 84.4      & 85.9      & 74.2      & 82.9     & 80.9       & 64.6       & 72.0      & 87.4      \\
                     & EfficientNetV2-SA    & 84.4      & 77.1      & 80.6      & 85.4     & 90.3       & 68.8       & 67.2      & 94.9      \\
400X                 & EfficientNetV2-S     & 67.1      & 82.9      & 61.8      & 94.8     & 93.6       & 57.5       & 78.0      & 72.5      \\
                     & EfficientNetV2-SA    & 74.2      & 83.4      & 70.5      & 84.2     & 95.3       & 60.1       & 76.1      & 67.5      \\ \hline
\end{tabular}}
\end{center}
\end{table*}

\subsection{Classification Accuracy}
The experiment mainly focuses on the recognition accuracy of the classifier.For multi-class problems, let K be the number of categories; TP, FP, TN, FN respectively represent the number of True Positive, False Positive, True Negative, and False Negative, and the subscript $\mathit i$ indicates which category it belongs to.

\begin{equation}
Accuracy=\frac{\sum_{i=1}^{K}TP_{i}}{\sum_{i=1}^{K}(TP_{i}+FN_{i})}
\end{equation}

Table VI shows the recognition rate performance of all the classification algorithms. This result indicates that our EfficientNetV2-SA network performs better than traditional machine learning classifiers on the task of classifying breast cancer images on all factors. The result of EfficientNetV2-S is closer to that of ResNet-50. However, EfficientNetV2-SA show significant improvements on the basis of EfficientNetV2-S, which are 6.4\%, 0.4\%, 1.4\% and 0.5\% on 40X, 100X, 200X and 400X datasets respectively.

We also test our model's accuracy on each single type breast cancer and compare it with EfficientNetV2-S. This can help us to figure out the performance of ACON-C in the case of uneven distribution of training data sets. Table VII shows the result of comparison. In these 32 classification tests, EfficientNetV2-SA lead in 23 tests. The highest improvement appears in the ductal carcinoma (DC) classification at 200X factor, which is 11.6\%. However, EfficientNetV2-SA's performances of 200X and 400X fluctuate and lag behind its opponents in some classes.

\subsection{Further Analyses}
Overall, the performance of the M-Chatbot is acceptable under the evaluation of the above technical metrics. Most of the users gave positive feedback that talking to the M-ChatBot can relieve them from the anxiety of potential health problems. However, the M-ChatBot is not always giving the desired response since we could not get enough medical corpus and consultation dialogues for chatbot training. For future improvement of the M-ChatBot, we could obtain more professional medical training data so that it can answer user’s questions better and more comprehensively.

Meanwhile, the image classification experimental results indicate that ACON-C activation in top layer can effectively improve the results of transfer learning and the classification performance of our EfficientNetV2-SA model is the best among several methods on most of the BreaKHis datasets at the same time. It also proves that this model can be applied in the medical pre-diagnosis system. Although recognition accuracy of our model on image data sets with small magnifications, it still has great potential for solving histopathological image classification tasks in the medical pre-diagnosis system. The performance of EfficientNetV2-SA may be better with the help of image pre-processing.

\section{Conclusion}
We study the combination of generative chatbot and pathological image classification, which is a potential medical diagnosis and treatment scenario. In this paper, we find out that ACON-C activation can be a facilitator of transfer learning on BreakHis dataset. Moreover, we proposes a brand-new diagnosis system, which can be divided into two components: a chatbot based on matching algorithm and a breast cancer pathological image classifier based on EfficientNetV2-SA. With this system, patients can complete initial autonomous diagnosis of breast cancer. In future, we will focus on improving the performance of this system in the corresponding latency of the dialogue and the image segmentation function.

\end{document}